\documentclass[letterpaper, 10 pt, , journal, twoside]{IEEEtran}  

\IEEEoverridecommandlockouts                              

\usepackage{xcolor}
\usepackage{float}
\usepackage{wrapfig}
\usepackage{amsmath}
\usepackage{booktabs}
\usepackage{multirow}
\usepackage{diagbox}
\usepackage{graphicx}
\usepackage{url}
\usepackage{caption}
\usepackage{subcaption}
\usepackage[ruled,vlined,noend,titlenotnumbered]{algorithm2e}
\usepackage{bbm}
\usepackage{amssymb}
\usepackage{siunitx}
\usepackage{wrapfig}
\usepackage{algorithmic}
\DeclareMathOperator*{\argmin}{arg\,min}

\DeclareMathOperator*{\z}{\mathbf{z}}

\begin{document}

\title{\LARGE \bf
Planning in Learned Latent Action Spaces for Generalizable Legged Locomotion
}

\author{Tianyu Li$^{1}$, Roberto Calandra$^{1}$, Deepak Pathak$^{2}$, Yuandong Tian$^{1}$, Franziska Meier$^{1}$, Akshara Rai$^{1}$
\thanks{ Manuscript received: Oct, 15, 2020; Revised: Dec, 17, 2020; Accepted: Feb, 17, 2021.}
\thanks{This paper was recommended for publication by
Editor Abderrahmane Kheddar upon evaluation of the Associate Editor and Reviewers’ comments. }
\thanks{$^{1}$FAIR, Menlo Park, CA, 94025, USA, 
        {\tt\small \{tianyul,rcalandra, yuandong,fmeier,akshararai\}@fb.com}}%
\thanks{$^{2}$ Robotics Institute, Carnegie Mellon University,
        Pittsburgh, PA, 15213, USA,
        {\tt\small dpathak@cs.cmu.edu}}%
        
\thanks{Supplementary Video Link: www.youtube.com/watch?v=yopO26MF-HU}
\thanks{Digital Object Identifier (DOI): see top of this page.}
}

\markboth{IEEE Robotics and Automation Letters. Preprint Version. Accepted Feb, 2021}
{Li \MakeLowercase{\textit{et al.}}: Planning in Learned Latent Action Spaces for Generalizable Legged Locomotion}

\maketitle

\begin{abstract}
Hierarchical learning has been successful at learning generalizable locomotion skills on walking robots in a sample-efficient manner. However, the low-dimensional ``latent'' action used to communicate between two layers of the hierarchy is typically user-designed. In this work, we present a fully-learned hierarchical framework, that is capable of jointly learning the low-level controller and the high-level latent action space. Once this latent space is learned, we plan over continuous latent actions in a model-predictive control fashion, using a learned high-level dynamics model. This framework generalizes to multiple robots, and we present results on a Daisy hexapod simulation, A1 quadruped simulation, and Daisy robot hardware. We compare a range of learned hierarchical approaches from literature, and show that our framework outperforms baselines on multiple tasks and two simulations. In addition to learning approaches, we also compare to inverse-kinematics (IK) acting on desired robot motion, and show that our fully-learned framework outperforms IK in adverse settings on both A1 and Daisy simulations. On hardware, we show the Daisy hexapod achieve multiple locomotion tasks, in an unstructured outdoor setting, with only 2000 hardware samples, reinforcing the robustness and sample-efficiency of our approach. 
\end{abstract}


\IEEEpeerreviewmaketitle

\section{Introduction}
Traditional control techniques used in legged locomotion, like inverse dynamics make assumptions about the dynamics of the robot, and can lead to poor performance when these assumptions are violated. In contrast, learning-based approaches do not make strict assumptions about dynamics, but are expensive to train. 
%
%
%
Learning locomotion skills can be made scalable to real robots by leveraging a two-layer hierarchical control structure~\cite{brain,us,deepgait}.
Typically in hierarchical control literature, the action space used by the high-level controller to interact with the low-level controller is user-defined~\cite{brain,us,deepgait,openai}.
Such a user-defined action space can potentially be too restrictive for some tasks. For example, \cite{us, openai} constrain the latent space to choose from a pre-defined set of primitives, limiting performance to the quality and number of primitives; \cite{deepgait} constrain the latent space to a footstep pattern, and learn a conservative walking pattern, while a different gait might move faster.

\begin{figure}[h]
    \centering
        \includegraphics[width=0.45\textwidth]{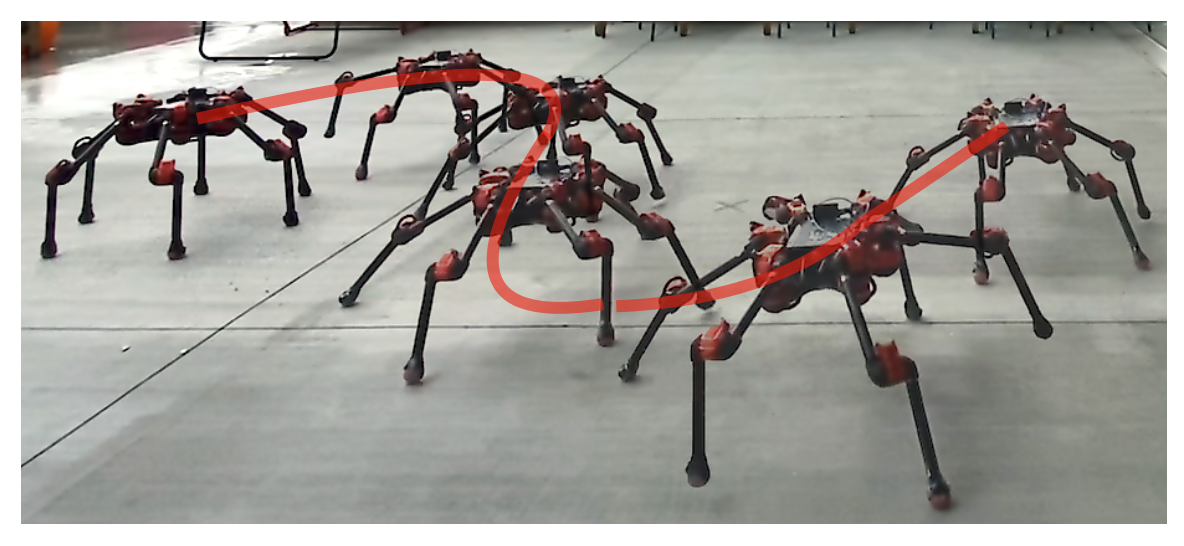}
        \caption{\small Daisy Hexapod tracking a desired trajectory in an outdoor unstructured environment using our approach, starting with only 2000 samples on hardware.}
        \label{fig:Daisy Hardware}
    \label{fig:latent_plot}
    \vspace{-0.5cm}
\end{figure}

\begin{figure*}
    \centering
        \includegraphics[width=0.95\textwidth]{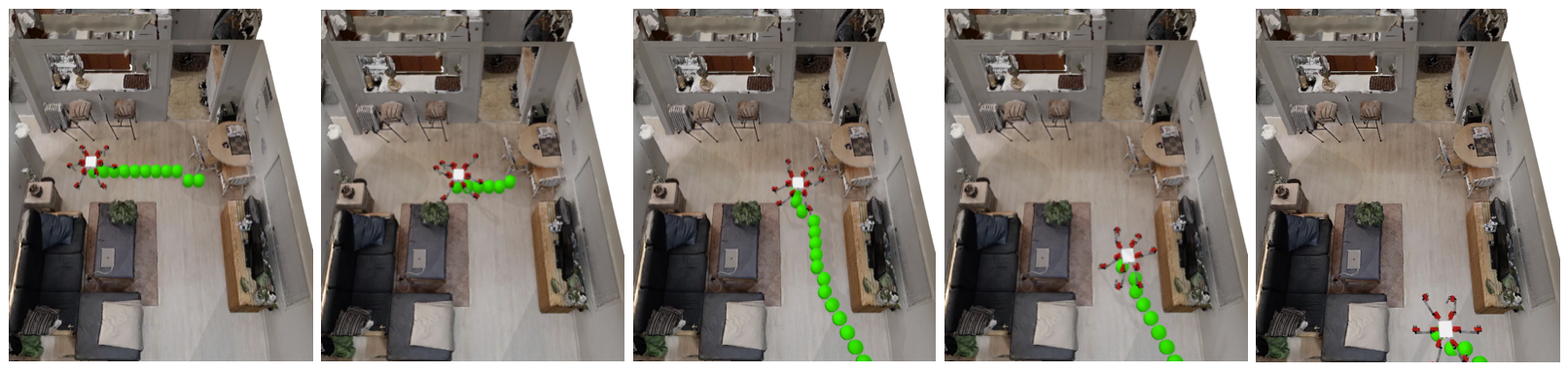}
        \vspace{-7pt}
    \caption{ \small Our method can directly be used with waypoint navigation methods to reach a desired goal. Here, our test platform `Daisy' navigates a cluttered photo-realistic indoor simulation environment in iGibson \cite{xia2020interactive} by following a collision-free path to goal. Our approach can follow the desired path, despite disturbances such as slipping and collisions with the environment.}
            \label{fig:Gibson}
    \vspace{-0.35cm}
\end{figure*}

In this work, we introduce a generalizable, fully-learned, hierarchical control framework that eliminates the need for pre-defined action spaces for the high-level controller. We start by jointly learning a low-level policy and a high-level latent action space using imitation learning to reproduce a set of experts in simulation. This step transforms the initially discrete space of expert primitives into a continuous space, allowing us to go beyond a finite number of primitives, while learning a suitable high-level latent action space. For the high level controller, we propose a model-based planner, and plan a sequence of learned latent actions to achieve a desired goal. We learn a `coarse' dynamics model over one cycle of the low-level policy given a latent action input, and use it for model-predictive control (MPC). This results in reactive planning in learned latent action spaces, allowing continuous modulation of robot motion to achieve changing targets, respond to disturbances, and generalize to new tasks.

Using the simulation of a Daisy hexapod and A1 quadruped (Figure \ref{fig:experiment_environment}), we compare our approach against different learned hierarchical approaches from literature and find that our approach outperforms baselines in all evaluation tasks, while being more sample-efficient. This includes a comparison between MPC on our learned latent action space and MPC on a library of experts that were used for learning our low-level, signifying the importance of jointly learning a high-level latent action space and a low-level policy. %
Furthermore, we compare our method with an informed baseline of inverse-kinematics (IK) acting on desired center of mass (CoM) motion. The IK baseline can be seen as the best-case low-level policy for a fixed action space, making it a highly competitive baseline rarely addressed in hierarchical learning literature. Our approach does not require any prior knowledge about the robot, unlike IK, but performs comparably to IK in normal conditions, and outperforms it in adverse settings. This experiment strongly reinforces the advantages of learning a high-level latent action space. Finally, we demonstrate that our approach can be used to solve complex locomotion tasks in a sample-efficient manner in the real world with a Daisy hexapod robot, with only 2000 hardware samples. 
Together, our experiments establish that our proposed framework generalizes to hardware, multiple robot simulations, and multiple tasks with no re-training per task, significantly improving the state-of-the-art of learned hierarchical controllers for locomotion. 

The main contributions of this work are: 1) Present a sample-efficient fully-learned hierarchical framework for locomotion and deploy it on a legged-robot. 2) Demonstrate a model-based planner to plan over learned high-level latent actions. 3) Extensively compare and analyze different learned as well as traditional hierarchical control schemes from literature. To the best of our knowledge, this is the first fully-learned model-based hierarchical framework demonstrated on a robot hardware. We present experiments on multiple robot simulations, further reinforcing the significance of our results. The generalizability and sample-efficiency of our approach makes it suitable for solving long-horizon legged locomotion problems, such as indoor navigation (Figure \ref{fig:Gibson}).

\section{Background and Related Work}
\paragraph{Model predictive control}
We consider a Markov Decision Process with actions $\mathbf{a}$ and states $\mathbf{s}$, cost~$c(\mathbf{s}_t,\mathbf{a}_t)$  and transition dynamics $\mathbf{s}_{t+1} = f_{dyn}(\mathbf{s}_t, \mathbf{a}_t)$. 
The objective of model-predictive control (MPC) is to minimize the long term cost of a trajectory $\tau$: $J = \sum_{t=0}^T \mathbb{E}_\tau [c(\mathbf{s}_t,\mathbf{a}_t)] $ 
with respect to the actions $\mathbf{a}_{0:T}$, given the dynamics $f_{dyn}(\mathbf{s}_t, \mathbf{a}_t)$, starting from an initial state $\mathbf{s}_0$. 
MPC plans a sequence of actions over a horizon $H$ $\mathbf{a}_{0:H} = \argmin_{\mathbf{a}_{0:H}} \sum_{h=0}^H c(\mathbf{s}_h, \mathbf{a}_h)$, given $\mathbf{s}_{h+1} = f_{dyn}(\mathbf{s}_h, \mathbf{a}_h)$ and $\mathbf{s}_0 = \mathbf{s}_{curr}$, starting from current state $\mathbf{s}_{curr}$.
The first action $a_0$ is applied, and the process repeats, starting with the new current state. MPC has been applied to locomotion control in \cite{koenemann2015whole}, \cite{mason2018mpc}, \cite{herdt2010online}.
In this work, we learn the transition dynamics $f_{dyn}$ from data, over a temporally-extended action sequence $\mathbf{a}_{1:N}$ of length $N$, i.e., $\mathbf{s}_{t+N} = f_{dyn}(\mathbf{s}_t, \mathbf{a}_{t:t+N})$. Similar dynamics models are used in \cite{atkeson} over Poincare sections of a bipedal gait. 

\paragraph{Learning for locomotion}
Classical locomotion controllers like~\cite{drc} require known dynamics, while machine learning approaches can learn locomotion without dynamics~\cite{sac}. Successful learning for locomotion has been demonstrated on robots using dynamics randomization~\cite{tan2018sim}, and other sim2real approaches like domain-specific features~\cite{rai, cully, yang2018learning}. \cite{peng, li2019using} iteratively learn a low-level policy and a latent input in simulation, and optimize the latent input on hardware using Bayesian optimization, or other sampling-based optimization approaches. While these approaches can learn to walk, they cannot directly learn skills that can generalize to new tasks, like multiple goals and different desired velocities. 
%
%
In contrast, hierarchical decomposition of control holds the promise of solving complex locomotion tasks in a general manner. For instance, \cite{drc,kalakrishnan2011learning,feng} decompose the problem of controlling a humanoid robot to center of mass planning, followed by model-based controllers. Also for learned controllers, \cite{brain, us, deepgait, openai} demonstrate the efficacy of hierarchical structure for solving locomotion tasks. \cite{openai} jointly learn a discrete low-level policy and a switching high-level policy in simulation. \cite{deepgait} use a learned high-level controller to decide a footstep location for a learned low-level policy in simulation. On hardware, \cite{us} use a high-level controller to sequence a set of pre-learned primitives on a hexapod robot and \cite{brain} use the high-level policy to define sub-goals for the low-level policy for a quadruped. However, these works assume a known high-level action space that communicates between the two levels of the hierarchy -- either as footsteps, or a discrete set of primitives. This can be restrictive if the pre-defined action space is not complex enough to represent a task. In contrast, we present a framework that can learn a continuous space of low-level primitives along with a learned high-level action space, and combine this with MPC to achieve multiple locomotion tasks.

\paragraph{Latent space learning for control}
There has been a lot of interest in the robotics community in learning latent representations of high-dimensional state like images, and using them for control, such as \cite{merel2018neural} \cite{watter2015embed}, \cite{banijamali2017robust}, \cite{finn2016deep}  and \cite{zhang2018solar}. 
However, there are fewer works that deal with latent actions, or latent inputs to a policy (without transition dynamics). Most closely related to our work is the work by \cite{mso}, \cite{yu} and \cite{peng}. 
These works iteratively learn a policy and a latent input to the policy on a large range of environments in simulation, and fine-tune the latent input on hardware. 
In contrast to these approaches, we jointly learn a high-level latent action space, and a low-level policy in a hierarchical setting. We also learn a dynamics model over this learned latent space and use MPC to plan over continuous latent actions, allowing us to change the latent action input to our low-level policy on the fly. As a result, each hardware trajectory in our approach can have different latent actions per step, as needed by the task, while the latent input is kept fixed in \cite{mso,yu, peng} for each hardware rollout. This allows us to respond to online disturbances and reach changing targets.

\section{Planning in a learned latent action space}
\begin{figure}
    \centering
    \includegraphics[width=\linewidth]{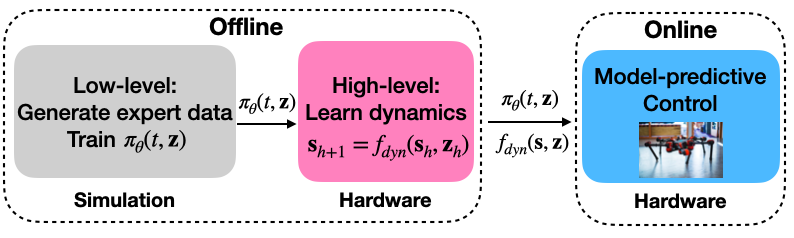}
    \caption{\small Flowchart explaining the learning pipeline of our proposed approach. (1) We learn a low-level policy $\pi_\theta(t, \z)$ from expert demonstration in simulation, where $t$ is the phase of the movement, and $\z$ is the latent action. (2) We learn the high-level dynamics $s_{h+1} = f_{dyn}(s_h, \z_h)$ that predicts the state after a latent action $\z_h$ is executed. (3) With $\pi_\theta$ and $f_{dyn}$, we use MPC on the high-level for planning in learned latent space.}
    \label{fig:learning_framework}
    \vspace{-0.35cm}
\end{figure}

Our hierarchical learning framework, illustrated in Figure~\ref{fig:learning_framework}, consists of three steps: 1) Jointly optimize a low-level policy $\pi_\theta$, and a latent action $\z$. 2) Learn a dynamics model given the learned low-level policy $\pi_\theta(\cdot,\z)$ over randomly sampled $\z$. 3) Plan over $\z$ using MPC and learned dynamics. This method can generalize to multiple robots and locomotion tasks while remaining sample-efficient, without re-training per task. 

\subsection{Hierarchical planning for locomotion}
We divide the control of legged robots into a low-level policy and a high-level model-based planner. The low-level policy $\pi_\theta$ runs at a frequency of $100$Hz and its behaviour is modulated by a latent input $\z$ from the high-level planner, which runs at about $1$Hz. Different $\z$ result in different behaviors on the robot. One popular example of this architecture is designing the low-level policy with inverse kinematics, and choosing high-level actions as CoM velocity. In contrast, we present a framework to jointly learn the  low-level policy, and latent action space, eliminating the need for user-defined action spaces. We also compare our method to the popular inverse kinematics setting in Section~\ref{sec:exps}. 

      \begin{algorithm}[t]
        \caption{Model-based planning on learned latent actions}
        \begin{algorithmic}

            \STATE Given $G$ expert demonstrations $\pi^{exp}_{1:G}$, high-level cost $\text{c}_{hl}$, horizon $H$\\
            
            \STATE Randomly initialize $G$ latent actions $\mathbf{z}_{1:G}$, $\pi_\theta$\\
            \For{each gradient step}{
            \STATE Update $\mathbf{z}^*_g, \pi_\theta^* = \argmin_{z_g, \pi_\theta} \mathcal{L}(z_g, \theta)$, \\
            \STATE where $\mathcal{L} = \sum_{g=1}^G \|\pi_\theta(t, \mathbf{z}_g) - \pi^{exp}_g(t)\|^2$
            }
            \For{each dynamics learning step}{
                \STATE Sample $\mathbf{z} \thicksim \text{unif}[\z_{\text{min}}, \z_{\text{min}}]$\\ 
                \While{$t \leq N$}{
                    \STATE $\mathbf{q}_{des}(t) = \pi_\theta^*(t, \mathbf{z})$ \\
              }
               
              \STATE $D \leftarrow D \cup \{(\mathbf{s}_0, \mathbf{z}, \mathbf{s}_{N})\}$\\
              \STATE Update dynamics model  $\mathbf{s}_{N} = f_{dyn}(\mathbf{s}_0,\mathbf{z})$ \\
              \STATE $\mathbf{s}_0 \leftarrow \mathbf{s}_{N}$\\
            }
            
            \For{each planning step}{
                \STATE $\mathbf{z}_{1:H} = \arg \min \text{c}_{hl}(\mathbf{s}_0, \mathbf{z}_{1:H})$ \\
                \STATE Apply latent action $\mathbf{z}_1$, measure $\mathbf{s}_N$ \\
                \STATE $\mathbf{s}_0 \leftarrow \mathbf{s}_N$
            }
        \end{algorithmic}
        \label{algo:HL algo}
        \end{algorithm}

\subsubsection{Low-level policy}
Our low-level policy is parametrized by a neural network that takes a phase variable $t \in (0,1]$ as input, along with a latent action $\z$ from the high-level planner.
$\mathbf{q}_{des,t} = \pi_\theta(t, \mathbf{z}) , \quad t = \frac{n}{N}\,$
where $n$ is the current time step, which goes up to the primitive length $N$, at which point it is reset to $0$ again. All primitives are assumed to be the same pre-determined length $N$. $\mathbf{q}_{des,t}$ are the desired joint-angles sent to the robot. 
The phase $t$ linearly increases to $N$, and ensures a cyclic nature of the low-level policy. For a fixed $\z$, the policy generates the same joint angle pattern every $N$ time steps in a periodic manner. 

\subsubsection{High-level planner}
\label{sec:high_level}
We use a model-based high-level planner that plans a latent action sequence $\z^{1:H}$ for a planning horizon $H$ using MPC. The dynamics used in this planning are learned over temporally-extended low-level action sequences, rather than per time-step transitions. 
Starting from the current center of mass (CoM) position and orientation $s_{curr} = (\mathbf{x}_{com}, \mathbf{\theta}_{com})$, our high level planner does a search over the possible sequences of latent actions to find the optimal sequence over a horizon $H$: 
\vspace{-0.25cm}
\begin{align}
    \mathbf{z}^*_{1:H} = \argmin_{\mathbf{z}_{1:H}} \sum_{h=1}^H \text{c}_{hl}(\mathbf{s}_h, \mathbf{z}_h) \\
    \text{s.t.} \quad \mathbf{s}_{h+1} = f_{dyn}(\mathbf{s}_h, \mathbf{z}_h), \quad  \mathbf{s}_{0}=\mathbf{s}_{curr}
    \vspace{-0.15cm}
    \label{eq:mpc}
\end{align}
$\text{c}_{hl}$ is the high-level cost function; $h$ is a robot step, consisting of $N$ time steps. The first action $\z^*_1$ is applied on the robot, and the optimization is repeated, starting from the new state. We optimize Equation \ref{eq:mpc} using random shooting with 8000 samples, over a horizon of $H=1$ and pick the best action. The cost is designed by the user depending on the task at hand, described in Section \ref{sec:exp_setting}. Using a learned dynamics model, MPC can generalize to multiple tasks without any re-training, hence significantly improving the sample-efficiency.
\begin{figure}[t]
    \centering
    \vspace{-0.25cm}
    \includegraphics[width=0.245\textwidth]{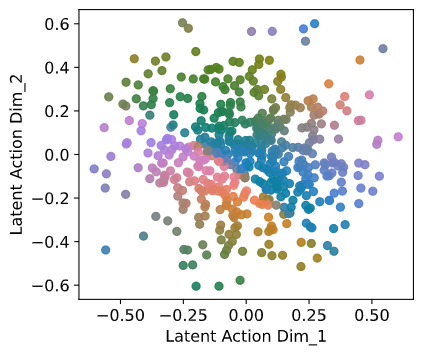}
    \includegraphics[width=0.234\textwidth]{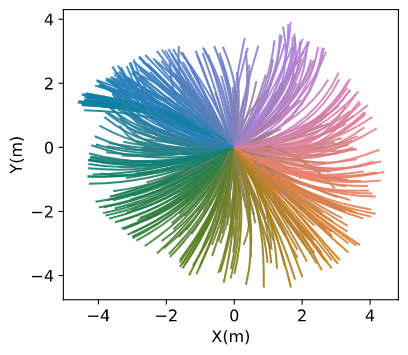}
    \caption{\small CoM Trajectory in XY plane caused by sampling in a 2-dimensional latent space. Close-by samples in the latent space lead to continuously varying CoM motion, showing a structured and stable learned latent space.}
    \label{fig:latent_visualization}
    \vspace{-0.5 cm}
\end{figure}
\subsection{Jointly learning low-level policy and latent actions}
We propose a framework for jointly learning a low-level policy and high-level latent actions, starting with $G$ expert demonstrations. 
We use supervised learning to jointly optimize a low-level policy $\pi_\theta$ and a latent action $\z$ input into $\pi_\theta$ that leads to a desired expert behavior $\pi^{exp}$.  Specifically, we optimize $\theta$ and one latent action per expert $\z_g$, such that the optimal policy $\pi_{\theta^*}(\cdot, \z^*_g)$ matches the expert $g$: 
%
\begin{equation}
    \pi_{\theta^*}(t, \mathbf{z}^*_g) = \pi^{exp}_g(t) \quad \text{for} \quad g = 1 \dots G
\end{equation}
This ensures that, if trained properly, $\pi_{\theta^*}$ is capable of generating at least $G$ experts, given different $\z^*_g$ per expert. This can be seen as a bottle-neck formulation where we want to jointly learn a policy parametrization $\theta^*$ and high-level latent action space $\z$ such that a large variety of gaits can be produced by the policy $\pi_{\theta^*}$ for different inputs $\z$. The loss for each gait $g$ becomes $\mathcal{L}_g = \|\pi_\theta(t, \mathbf{z}_g) - \pi^{exp}_g(t)\|^2 $. $\theta$ and $\z_{1:G}$ can now be optimized together to reduce the total loss
\vspace{-0.15cm}
\begin{equation}
    \theta^*, \mathbf{z}^*_{1:G} = \argmin_{\theta, \mathbf{z}_{1:G}} \sum_{g=1}^G \|\pi_\theta(t, \mathbf{z}_g) - \pi^{exp}_g(t)\|\,^2.
\end{equation}
%
%
%
%
%
The partial derivatives of the total loss $\mathcal{L}$ with respect to latent variable $\z_g$ and the policy weights $\theta$ are:
\vspace{-0.1cm}
\begin{align}
     \frac{\partial \mathcal{L}}{\partial \theta} = \frac{\partial }{\partial \theta}\sum_{g=1}^G \|\pi_\theta(t, \mathbf{z}_g) - \pi^{exp}_g(t)\|^2,\\
     \quad
     \frac{\partial \mathcal{L}}{\partial \mathbf{z}_g} = \frac{\partial }{\partial \mathbf{z}_g}\|\pi_\theta(t, \mathbf{z}_g) - \pi^{exp}_g(t)\| \,^2.
\end{align}
This naturally leads to a formulation where the update to each latent input $\z_g$ is only affected by the supervised learning loss of the expert $g$ that it is trying to imitate. On the other hand, updates to low-level policy parameters $\theta$ are optimized by reducing the loss over all $G$ experts. 
This is in contrast to \cite{mso, peng} that iteratively optimize the latent action and policy which can lead to unstable training. Moreover, unlike \cite{peng} we learn a unified low-level policy across all experts, and not separate policies for different experts.

The expert data for supervised learning is collected using a model-based hierarchical controller proposed in \cite{truong2020learning}. An expert-designed feedback law converts desired CoM velocities to footsteps, which are followed using a low-level inverse kinematics (IK) policy. We randomly sample 50 desired CoM velocities, and use the generated joint angle trajectories as the experts used in training the low-level policy $\pi_\theta$ and latent action space $\z$. Once $\pi_\theta$ is learned, we are no longer limited to the $G$ experts that were used during training. By sampling in the space of latent actions $\z$, we can generate new controllers that interpolate, and extrapolate from the training experts. As illustrated in Figure \ref{fig:latent_visualization}, different $\z$ lead to different CoM trajectories. Close-by samples in the latent action space usually lead to continuous behaviors in the CoM space, showing that our learned latent space is structured. 

\subsection{Learning center of mass dynamics}
Once the policy $\pi_\theta$ and latent space $\z$ is learned, we can plan over $\z$ to accomplish a variety of locomotion tasks, like reaching goals and trajectory tracking. To do this, we learn a temporally-extended dynamics model of the center of mass (CoM), learned by sampling a series of random latent actions $\z$ and measuring the resultant CoM state after one cycle of the low-level policy $\pi_\theta(\cdot, \z)$. 
This leads to a `coarse' dynamics model where the next state $s_{t+N} = f_{dyn}(s_t, \z_t)$ is the state after executing the low-level policy for $N$ time steps, starting from $s_t$ and using latent action $\z_t$.

In a similar spirit, previous works like \cite{drc, feng} have used approximate CoM dynamics models such as a linear inverted pendulum (LIPM). Unlike LIPM, our CoM dynamics model is learned directly from robot data, and hence adheres to the dynamics of the robot. For example, the dynamics learn that certain latent actions result in slipping on the robot, and hence lead to slower CoM movement. As a result, slipping and other unobserved factors are directly absorbed in our dynamics, eliminating the need for no-slip constraints like friction cones, as in \cite{drc, feng}. Moreover, we learn dynamics in a learned latent action space, and are not restricted to planning in a user-defined space like CoM velocity. 

Formally, we represent current state of the CoM by $s_{curr} = (x_{curr}, y_{curr}, \gamma_{curr}, \dot{x}_{curr}, \dot{y}_{curr})$, where $(x_{curr}, y_{curr})$ is the current horizontal position, $(\dot{x}_{curr}, \dot{y}_{curr})$ is the current horizontal velocity and $\gamma_{curr}$ is the current yaw. The learned dynamics represent a transition from the current state to the next state $s_{next} = (x_{next}, y_{next}, \gamma_{next}, \dot{x}_{next}, \dot{y}_{next})$, using the low-level controller $\pi_\theta$ with latent action $\z$. To simplify the learning and improve generalization, we learn a delta dynamics model in the current local CoM frame
\vspace{-0.15cm}
\begin{align}
    \Delta x, \Delta y, \Delta \gamma, \dot{x}_{next}, \dot{y}_{next} = f_{dyn}(\dot{x}_{curr}, \dot{y}_{curr}, z) 
\end{align}
where $\Delta x = x_{next} - x_{curr}$ , $\Delta y = y_{next} - y_{curr}$  and $\Delta \gamma ~=~\gamma_{next} - \gamma_{curr}$. The delta predictions and measurements of $s_{curr}$ can be used to predict the next state $s_{next}$.

A learned dynamics model for high-level planning has several advantages, over learning a model-free high-level policy: 1) It is much more sample-efficient  2) It generalizes to new tasks, and needs no additional re-training for new tasks. In our experiments in Section \ref{sec:exps}, we show that this model-based planner outperforms model-free approaches, like \cite{brain}, in sample-efficiency and overall performance.

We do not explicitly model the CoM dynamics in the vertical plane, or the roll and pitch of the robot, to improve sample-efficiency of dynamics learning. Implicitly, controllers with lower CoM height, or high roll and pitch lead to slower gaits, which is learned by our dynamics.

\section{Experiments}
\label{sec:exps}
We evaluate our proposed framework on two robot simulations (Daisy and A1) and one real robot (Daisy). We provide extensive comparisons to prior learning-based methods, as well as, a traditional control approach.
Our results demonstrate that our approach outperforms current state-of-the-art in learned hierarchical locomotion, generalizes to multiple robots and solves real-world locomotion tasks. 
\subsection{Experimental Setup}
\label{sec:exp_setting}
\begin{figure}[t]
    \centering
    \begin{subfigure}[b]{0.15\textwidth}
        \centering
        \includegraphics[width=0.95\textwidth]{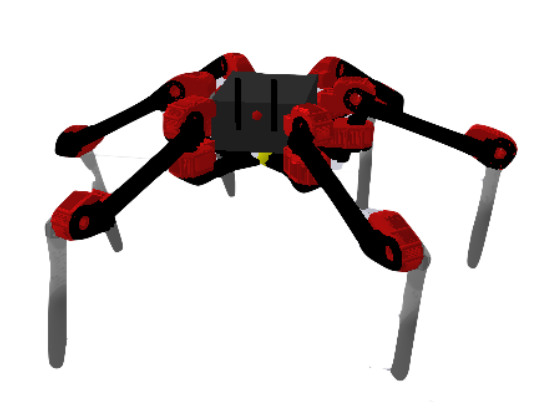}
        \caption{Daisy Simulation}
        \label{fig:Daisy Simulation}
    \end{subfigure}
    \hfill
    \begin{subfigure}[b]{0.15\textwidth}
        \centering
        \includegraphics[width=0.95\textwidth]{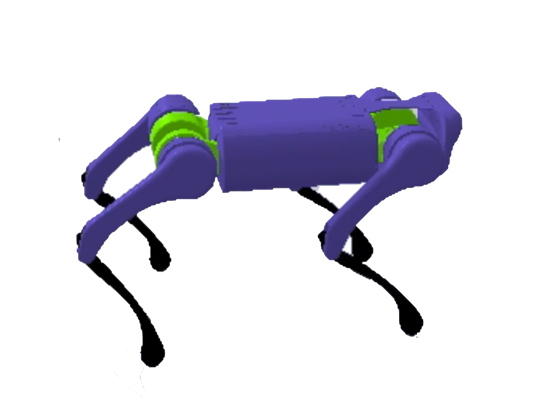}
        \caption{A1 Simulation}
        \label{fig:A1 Simulation}
    \end{subfigure}
    \hfill
    \begin{subfigure}[b]{0.15\textwidth}
        \centering
        \includegraphics[width=0.95\textwidth]{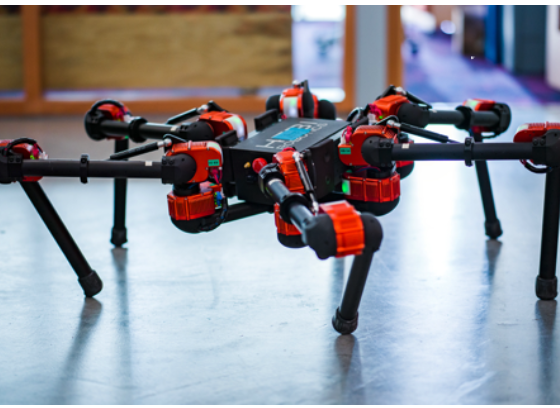}
        \caption{Daisy Hexapod}
        \label{fig:Daisy Hardware}
    \end{subfigure}
    \label{fig:latent_plot}
    \vspace{-0.05cm}
    \caption{\small Our experimental platforms: Daisy in Pybullet, A1 in Pybullet, and a real-world Daisy. }
    \vspace{-0.6cm}
    \label{fig:experiment_environment}
\end{figure}

We use Pybullet~\cite{pybullet} as our physics simulator, and build models for two robots: Daisy hexapod robot from Hebi robotics \cite{hebi} and A1 quadruped from Unitree Robotics \cite{unitree}, shown in Figure \ref{fig:Daisy Simulation} and \ref{fig:A1 Simulation} respectively.
Daisy is a 6-legged robot, with 3 motors in each leg, weighing \SI{21}{kg} and \SI{1.1}{m} by \SI{1.1}{m} wide in its nominal stance. A1 is a 4-legged robot, with 3 motors in each leg, weighing \SI{12}{kg} and \SI{0.5}{m} by \SI{0.3}{m} in dimensions. They have very different robot morphologies (hexapedal vs. quadrupedal) and significantly different mechanics and dynamics. 
On hardware, we use the Daisy robot with a Realsense camera \cite{realsense} to measure the position, orientation and velocity of the robot in an unstructured outdoor space. The position and velocity feedback gains for the base, shoulder and elbow were $[2.0,3.0,4.0]$ and $[0.2, 0.1, 0.15]$ for all legs in all hardware experiments.

To evaluate our method, we create a series of tasks that are relevant to real-world locomotion. 
%
%
%

\noindent \textbf{Velocity tracking:} This task measures the robot's ability to track a desired CoM velocity $\mathbf{v}_{tgt}$ and orientation $\gamma_{tgt}$, leading to cost $c_1 = w_1||\mathbf{v}_{tgt} -  \mathbf{v}_{curr} || + w_2 ||\gamma_{tgt} -  \gamma_{curr} ||$, where $[w_1, w_2] = [2, 1]$. The target velocity is varied throughout the experiment and the controller has to adapt to new targets. Target velocities were $[0.0,0.2]$m/s, $[0.2,0.0]$m/s, $[0.0,-0.2]$m/s, $[-0.2, 0.0]$m/s where each variable indicates desired velocity in $x$ and $y$ directions respectively.

\noindent \textbf{Goal reaching:} Some real-world locomotion tasks involve reaching a goal in space, either as a long-distance goal, or an intermediate waypoint. This tasks requires the robot to reach a range of CoM targets $s_{tgt} = (x_{tgt}, y_{tgt})$ and orientation $\gamma_{tgt}$ using cost $c_2 = w_1||s_{tgt} -  s_{curr} || + w_2||\gamma_{tgt} -  \gamma_{curr} ||$, where $[w_1, w_2] = [2, 1]$. Eight desired goals were uniformly distributed on a circle of radius 2m, with target orientation always pointing towards y axis.

\noindent \textbf{Trajectory tracking: } For navigating cluttered spaces with obstacles, or controlled environments, like a road, it is important to closely follow a CoM trajectory designed by a planner. We track an S-shaped desired trajectory consisting of target CoM positions $s_{tgt, t}$ and orientations $\gamma_{tgt, t}$ that change with time $t$, leading to cost $c_3 = w_1||s_{tgt, t} -  s_{curr} || + w_2||\gamma_{tgt, t} -  \gamma_{curr} ||$, where $[w_1, w_2] = [2, 1]$. 

This is the first detailed study of hierarchical learning for locomotion on a rich set of tasks, including demonstration on hardware. While \cite{deepgait, openai} have been shown to achieve interesting results in simulation, they have not been demonstrated on hardware, or extensively analyzed in simulation. On the other hand, \cite{brain, us} only show results on a subset of our designed tasks on hardware. Our analysis highlights the relative advantages and disadvantages of the hierarchical choices made in literature and helps make significant, scientific conclusions about the different approaches. By experimenting on two different robot designs, we further reinforce the statistical significance and generality of our results.

\begin{figure}[t]
    \centering
    \includegraphics[width=0.45\textwidth]{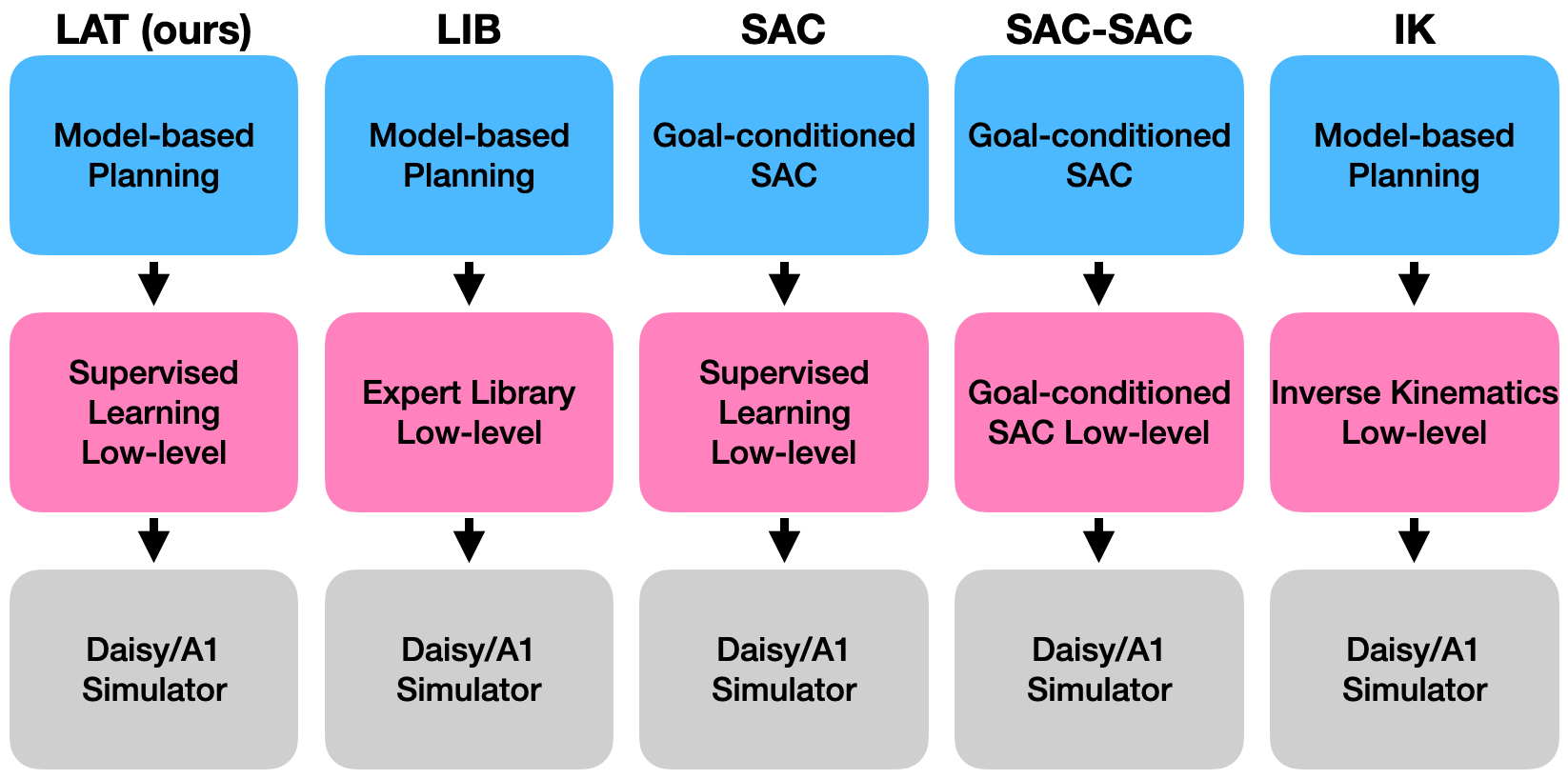}
    \caption{\small Flowchart of the different approaches from literature that we compare our approach \textbf{(LAT)} against. \textbf{LAT}, \textbf{LIB} and \textbf{IK} use a learned high-level CoM dynamics model with a learned policy, an expert-library, and an IK-based low-level policy, respectively. \textbf{SAC} and \textbf{SAC-SAC} are model-free high-level approaches, with low-level policies learned using supervised and reinforcement learning respectively. All neural network policies and dynamics models have 2 hidden layers with 512 nodes and ReLU activation.}
    \label{fig:baseline methods}
    \vspace{-0.6cm}
\end{figure}
\begin{figure*}[t]
    \centering
    \begin{subfigure}[b]{0.24\textwidth}
        \centering
        \includegraphics[width=0.995\textwidth]{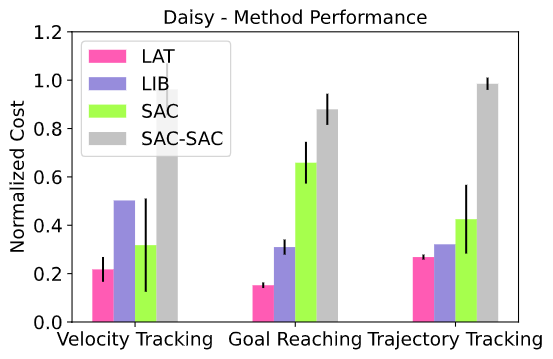}
        \vspace{-0.553cm}
        \caption{\scriptsize  Baseline Comparison - Daisy}
        \label{fig:baseline result}
    \end{subfigure}
    \begin{subfigure}[b]{0.24\textwidth}
        \centering
        \includegraphics[width=0.995\textwidth]{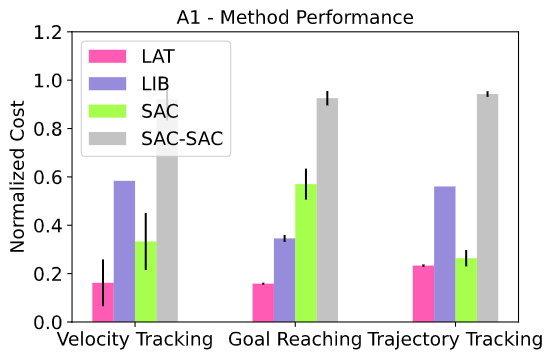}
        \vspace{-0.553cm}
        \caption{\scriptsize Baseline Comparison - A1}
        \label{fig:LAT_VS_IK}
    \end{subfigure}
    \begin{subfigure}[b]{0.24\textwidth}
        \centering
        \includegraphics[width=0.995\textwidth]{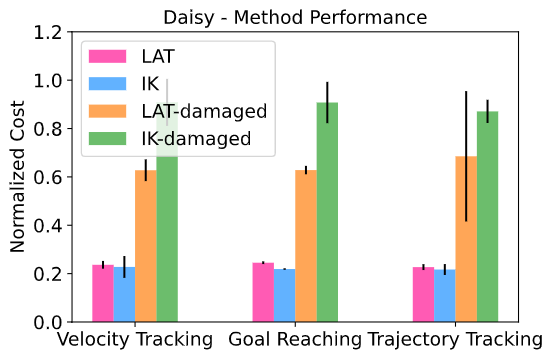}
        \vspace{-0.553cm}
        \caption{\scriptsize LAT \& IK Comparison - Daisy}
        \label{fig:daisy ik result}
    \end{subfigure}
    \begin{subfigure}[b]{0.24\textwidth}
        \centering
        \includegraphics[width=0.995\textwidth]{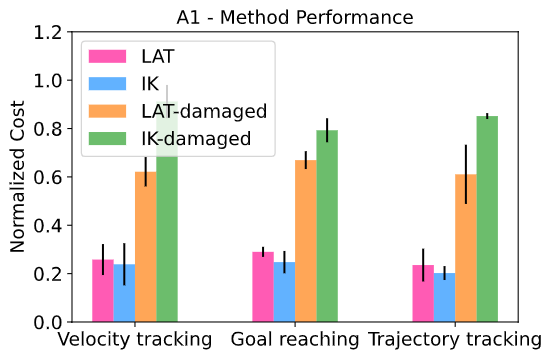}
        \vspace{-0.553cm}
        \caption{\scriptsize LAT \& IK Comparison - A1}
        \label{fig:a1 ik result}
    \end{subfigure}
    \caption{\small  (a), (b) Comparison of learned hierarchical approaches from Section~\ref{fig:baseline methods} on Daisy and A1 simulation. Our framework with a model-based high-level planner and supervised learning for low-level policy outperforms the baselines in all test tasks. (c), (d) Comparison with the IK baseline from Section \ref{sec:IK}. Our approach performs similar to IK in the normal setting, and outperforms it in adverse settings.}
        \label{fig:baseline comparison}

    \vspace{-0.5cm}
\end{figure*}
\subsection{Comparison with hierarchical baselines}
To compare our framework against these prior approaches, we create an ablation experiment in simulation. We characterize all hierarchical approaches by the choice of their low-level and high-level policies, and sweep through the different choices made in literature. Finally we test all the approaches on A1 and Daisy robot simulations and tasks described in Section \ref{sec:exp_setting}. Note that all model-free approaches had to be re-trained for new tasks, and hence, \textit{effectively take 3 times as much data} as model-based approaches.
The approaches compared are shown in Figure~\ref{fig:baseline methods} and can be broadly categorized into the following groups:

\noindent \textbf{A library of experts (LIB):} \cite{us, nachum2018data} use a library of low-level experts, and learn a high-level policy that chooses between the experts. We consider the same library of $50$ experts that was used for supervised learning of our low-level policy, and used model-based planning for the high-level. This comparison highlights the importance of a learned high-level action space versus a discrete set of expert primitives.

\noindent \textbf{Model-free high-level policy (SAC): } An alternative to model-based planning is to learn a model-free high-level policy, as in  \cite{brain}, \cite{nachum2018data}. Assuming the same learned low-level policy as our approach, we use goal-conditioned SAC \cite{sac} to learn a high-level policy. This setting is also similar to \cite{peng} and \cite{mso}, though these approaches cannot directly generalize to multiple targets. This motivates the importance of model-based planning in learned latent action spaces for generalization versus learning a model-free policy.

\noindent \textbf{Model-free learned low-level and high-level policy (SAC-SAC): }  \cite{brain} use model-free RL for training a low-level policy designed to reach sub-goals, and train a model-free high-level policy to define sub-goals for the pre-trained low-level policy. We use goal-conditioned SAC to learn both levels of this hierarchy, first training the low-level, then training the high-level, keeping the low-level fixed. This experiment compares our framework with model-free frameworks like \cite{brain, deepgait} on a wide-range of locomotion tasks. It highlights the importance of jointly learning a high-level action space with a low-level policy, followed by model-based planning.

The summary of these comparisons is shown in Figure~\ref{fig:baseline result} and  \ref{fig:LAT_VS_IK} for Daisy and A1 robot simulations. We used 10,000 samples for training the high-level dynamics and policies with an Adam optimizer, 1e-3 as learning rate and 512 as batch size. All model-free approaches are trained on half of the targets, and tested on all targets. For example, goal-conditioned SAC is trained to reach 4 goals, and tested on all 8. Both \textbf{SAC} and \textbf{SAC-SAC} are re-trained for new tasks, hence using 3 times as much data as our approach (\textbf{LAT}). \textbf{SAC} and \textbf{SAC-SAC} fail to generalize to new targets and perform worse than our approach on all tasks. On A1, \textbf{SAC} incurs a cost of $0.57 \pm 0.06$ at goal reaching, and \textbf{SAC-SAC} incurs a cost of $0.92 \pm 0.04$, much higher than \textbf{LAT} with $0.34 \pm 0.02$. This is not surprising as model-free approaches are known to be less sample-efficient and poorer at generalizing than model-based approaches \cite{pets}. \textbf{SAC-SAC} also has a user-defined high-level action space that performs poorly than the learned action space used in both \textbf{LAT} and \textbf{SAC}. This experiment reinforces our claim that model-based planning in learned action spaces improves performance and generalization.

The baseline \textbf{LIB} uses a discrete set of primitives with a model-based high-level controller, and generalizes well to goal reaching. For velocity tracking, however, the low-level primitives in \textbf{LIB} do not include experts that can reach the target velocity, and hence \textbf{LIB} incurs a high cost of $0.58 \pm 0.01$ on A1. On the other hand, \textbf{LAT} can plan in the learned latent action space and interpolate between the discrete set of primitives, incurring a much lower cost of $0.16 \pm 0.10$ on A1. This experiment proves our hypothesis that a jointly learned continuous high-level action space and low-level policy can outperform a discrete set of low-level primitives. Our approach combines the robustness of model-based planning with learned high-level and low-level controllers, and can successfully solve a large range of tasks sample-efficiently, without re-training per task. 


\subsection{Comparisons with a Inverse Kinematics in simulation}
\label{sec:IK}
Next, we present a comparison with a structured model-based hierarchical control scheme, where the high-level controller plans over desired CoM velocities, which are converted into desired footsteps using an expert-designed feedback law, and followed using IK, as proposed in \cite{truong2020learning}. Similar control schemes have been successful at solving a large range of locomotion tasks~\cite{drc}. This is an easy change for our setup, where the low-level policy $\pi_\theta$ is replaced by IK, and the desired CoM velocities works as an `oracle' high level latent space. 
A high-level model-based planner uses a learned CoM transition dynamics model, similar to Section \ref{sec:high_level}, with desired CoM velocities as the `latent' action. IK represents the best-case low-level policy for a fixed action space, and hence makes a highly challenging comparison for our approach. Unlike IK, our approach does not use kinematic knowledge of the robot, and can outperform IK when kinematic assumptions are violated. 
%
With this in mind, we run comparison experiments with IK in a normal setting, as well as, an adverse setting where two hind-legs are disabled by fixing their joint angles in simulation. The high-level CoM dynamics model is re-learned for both \textbf{LAT} and \textbf{IK} with broken legs, but the low-level policies are the same as those of the normal setting. The results are summarized in Figure \ref{fig:daisy ik result} and \ref{fig:a1 ik result} for Daisy and A1 simulations, and show that our fully-learned approach is close to IK in the normal setting. However, when the two hind-legs are disabled, IK's performance deteriorates significantly to $0.91 \pm 0.06$ at velocity tracking on A1, while our learned latent space compensates for this adverse setting, incurring a cost of $0.62 \pm 0.06$. This experiment highlights that a learned action space adds additional robustness to hierarchical control, including when the robot is damaged and other adverse scenarios.
%
\begin{figure}[h]
    \centering
    \includegraphics[width=0.4\textwidth]{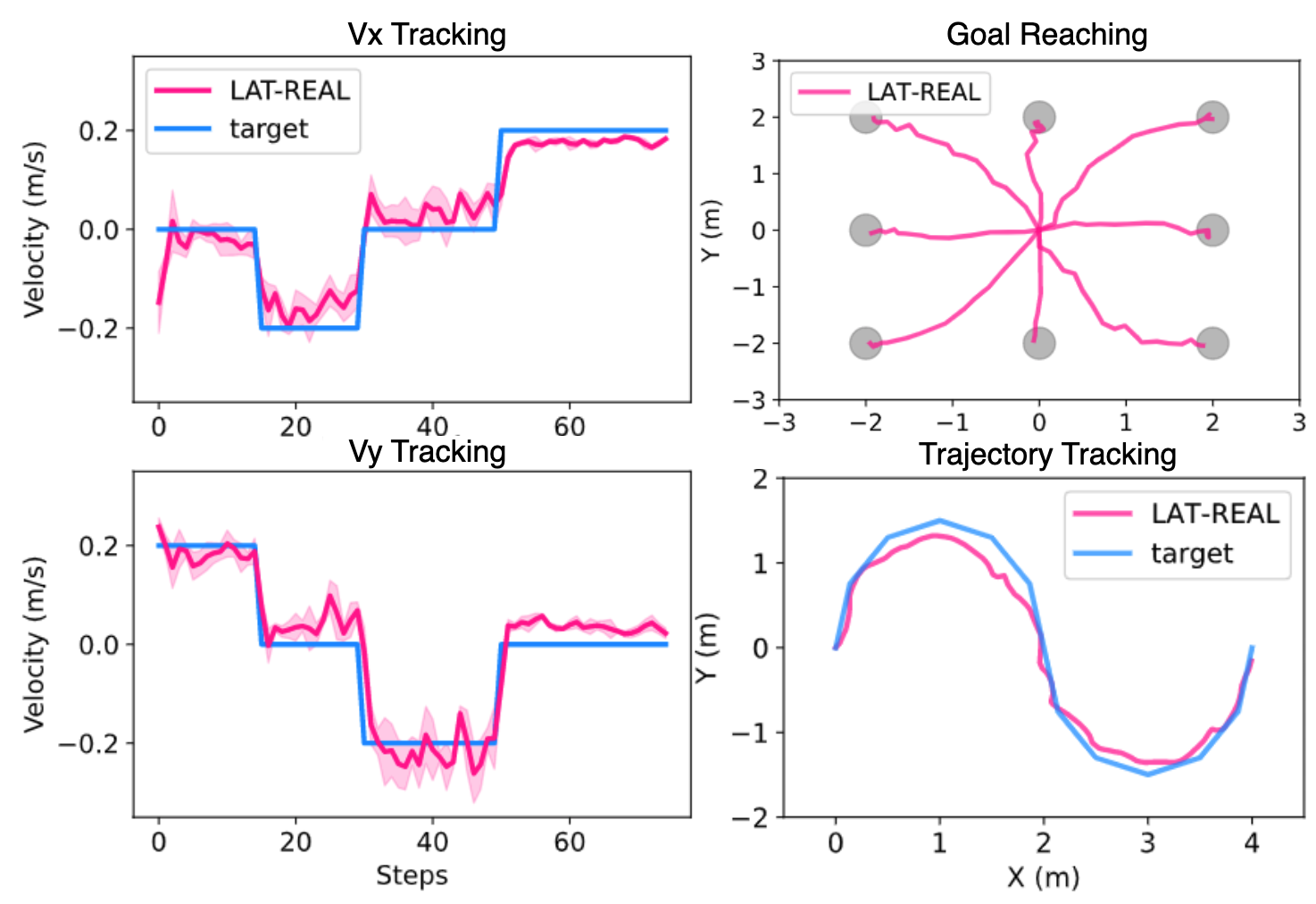}
    \caption{\small Performance in the real world on the Daisy hexapod over 3 hardware trials, with only 2000 robot samples. Velocity tracking (left), goal reaching (right-top), and trajectory tracking (right-bottom) on hardware. Our approach generalizes to hardware and achieves multiple multiple locomotion tasks sample-efficiently.}
    \label{fig:LAT_IK_hardware}
    \vspace{-0.6cm}
\end{figure}

\begin{figure*}[h]
    \centering
        \includegraphics[width=0.92\textwidth]{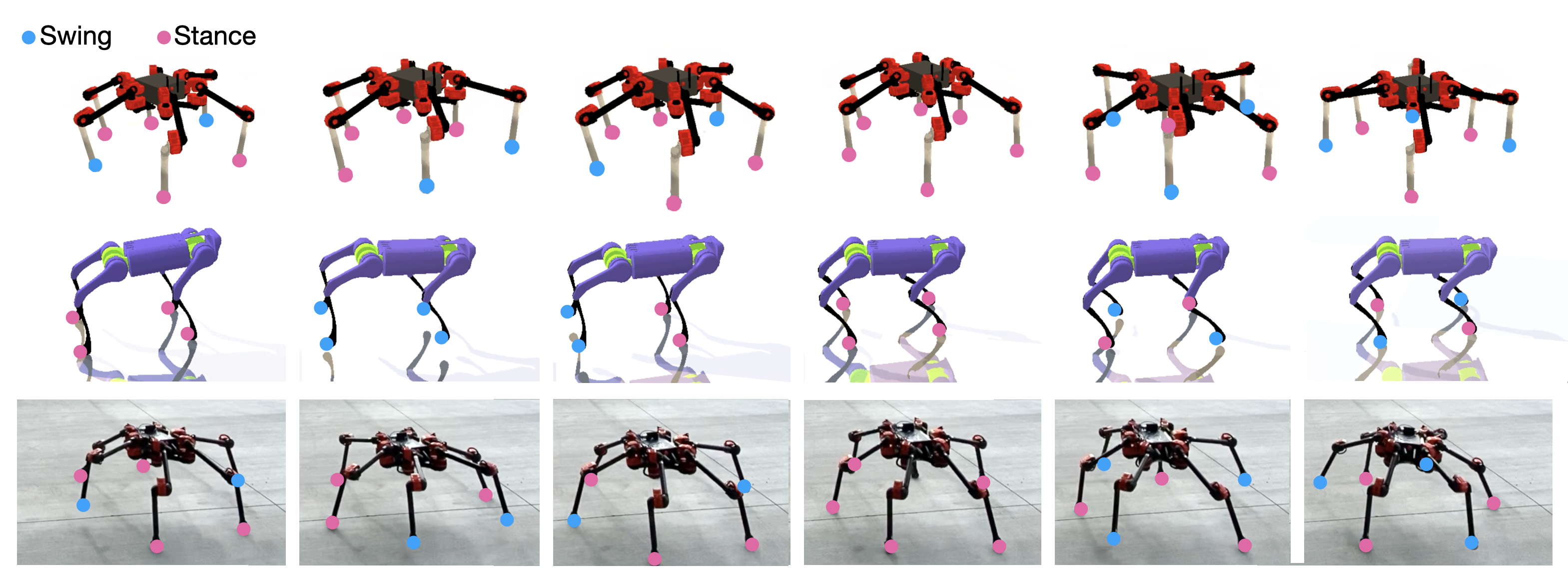}
        \vspace{-0.1cm}
    \caption{\small Gait transitions observed during planning in the learned latent action space. Pink indicates stance legs while blue indicates swing. Our learned latent space can induce gait transitions in the low-level if needed by the task.}
            \label{fig:a1_gait_transition}
    \vspace{-0.5cm}
\end{figure*}

\subsection{Hardware experiments on Daisy Hexapod}
\begin{figure}[h]
    \centering
    \begin{subfigure}[b]{0.235\textwidth}
        \centering
        \includegraphics[width=0.9\textwidth]{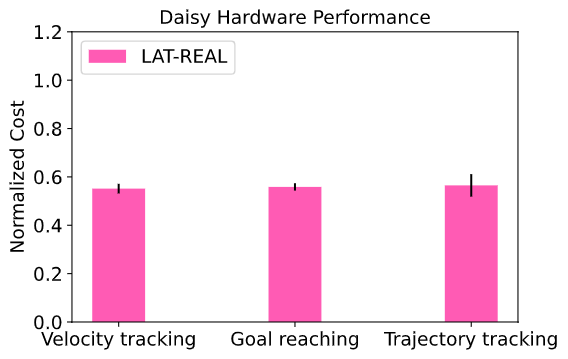}
        \vspace{-0.3cm}
        \caption{\scriptsize Daisy hardware performance.}
        \label{fig:daisy_hardware_experiments}
    \end{subfigure}
    \begin{subfigure}[b]{0.235\textwidth}
        \centering
        \includegraphics[width=0.9\textwidth]{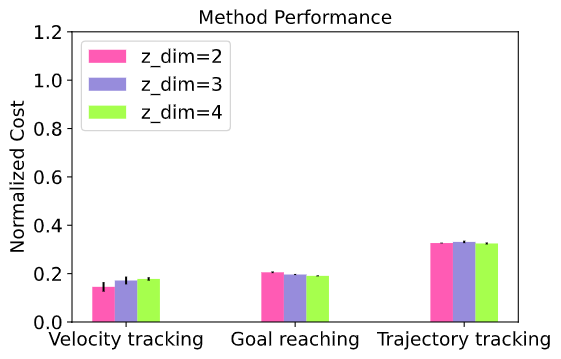}
        \vspace{-0.3cm}
        \caption{\scriptsize Ablation on latent space dimension}
        \label{fig:dim of z}
    \end{subfigure}
    \begin{subfigure}[b]{0.235\textwidth}
        \centering
        \includegraphics[width=0.9\textwidth]{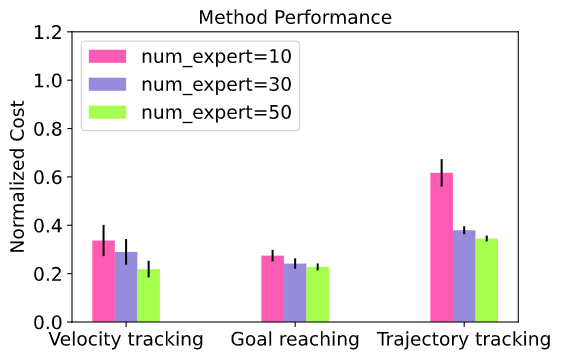}
        \vspace{-0.3cm}
        \caption{\scriptsize Ablation on number of experts}
        \label{fig:num_expert}
    \end{subfigure}
    \begin{subfigure}[b]{0.235\textwidth}
        \centering
        \includegraphics[width=0.9\textwidth]{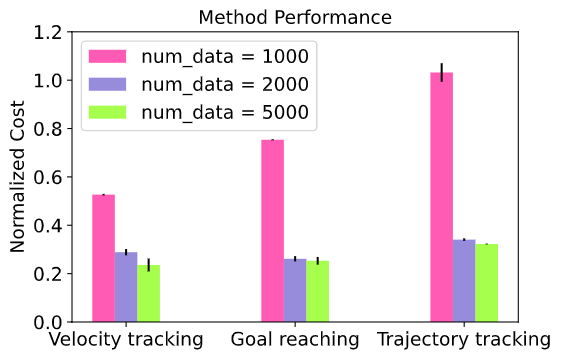}
        \vspace{-0.3cm}
        \caption{\scriptsize Ablation on dynamics learning}
        \label{fig:num_samples}
    \end{subfigure}
    \caption{\small (a) Hardware results on Daisy hexapod in the real-world. Daisy solves all test tasks despite noise like sensing errors, slipping, etc. (b)-(d) Ablation study on sensitivity to different hyperparameters. Our approach is not sensitive to hyperparameters once a good dynamics model and low-level policy has been learned.}
    \label{fig: ablation study}
    \vspace{-0.5cm}
\end{figure}

%
Transfer to hardware is straight-forward for our approach, once the latent action space and low-level policy have been learned in simulation. We keep the low-level policy and latent action space from simulation fixed on hardware, and learn the CoM dynamics by randomly sampling latent actions on the Daisy robot hardware (Figure \ref{fig:Daisy Hardware}). Once the dynamics models is learned, it is kept fixed, as we conduct hardware experiments, which span multiple days. Since our hardware experiments are conducted outdoors, in an uncontrolled environment, they naturally include noise due to tracking errors by the Realsense camera, slipping on gravel, joint tracking and ground height disturbances.
We use 2000 hardware samples for learning the dynamics model, and 20 simulation expert controllers for learning the low-level policy. The performance of our approach on hardware is shown in Figure \ref{fig:LAT_IK_hardware}. The behavior of the robot during these experiments can be seen in the supplementary video. 
For goal reaching, our approach could reach all goals in $16.0 \pm 1.4$ steps, averaged over 4 goals and 3 trials. The velocity tracking error on hardware was $0.56 \pm 0.015$ and trajectory tracking error was $0.57 \pm 0.05$ (Figure \ref{fig:daisy_hardware_experiments}).

Our approach was successfully able to track the desired trajectory, and reach desired goals on the Daisy hardware. Its performance was worse than in simulation because: 1) The number of samples used for learning dynamics on hardware were 1/5 of the samples used in simulation. 2) Learning dynamics on hardware was harder due to noise from poor joint tracking, slipping and noise in CoM estimation. 3) The robot was unable to achieve some high velocity targets, as the low-level controllers that could achieve high velocities in simulation did not transfer well to hardware.
The first two issues can be alleviated by using more hardware samples for learning the CoM dynamics model, or starting from a simulation dynamics model and fine-tuning. The third issue was caused because the simulation does not sufficiently capture the motor bandwidth of the robot, which can be solved by using dynamics-randomization, or higher-fidelity simulators. We leave this to future work.

\subsection{Ablation study of hyperparameters}

Lastly, we  conduct an ablation experiment where we change the dimension of our latent space, number of expert controllers and number of samples used in training the dynamics model in Daisy simulation. We vary the latent action dimensions to be 2, 3, 4, and find that there is no significant effect on the performance. Next, we use different numbers of expert controllers -- 10, 30, 50 experts -- for training the low level policy. As the number of experts increases the performance on each tasks improves, though the difference from 30 to 50 is minor. This shows that a low-level policy learned with 30 experts can generalize well, and perform similar to a low-level policy with 50 experts. Lastly, we vary the number of samples used for learning the high-level dynamics, using 1000, 2000 and 5000 datapoints. We observe a significant improvement in performance when the dynamics is learned from 2000 data points versus 1000, showing that a better dynamics model leads to better performance. However, there is no significant difference in performance between 2000 and 5000 data points, signifying that our approach is not sensitive to inaccurate dynamics model, and can correct for small dynamics errors. The result are summarized in Figure \ref{fig: ablation study}, and each test setting is averaged over 10 independent trials. These results further reinforce the robustness and sample-efficiency of our proposed approach.

\section{Discussion and Future Work}
%
\vspace{-0.1 cm}
\noindent \textbf{Gait transitions in learned latent space:} When the low-level policy is learned with experts spanning multiple gaits, the high-level latent action space can be used to transit between gaits, if needed by the task. We observe this on both robot simulations, as well as hardware, shown in Figure \ref{fig:a1_gait_transition}. Exploring gait transitions for achieving complex tasks is a promising extension to our current results that we aim to explore in the future.

\noindent \textbf{Diversity of expert gaits for low-level training:} For our hardware experiments, we used 20 experts from simulation to learn the low-level policy. The richness of the expert library directly affects the performance of the low-level controller, and since some of the low-level controllers did not transfer to hardware, the performance of our approach suffered. One of the challenges of our approach is to build a sufficiently diverse low-level policy, that can transfer to hardware. A combination of curiosity-driven learning, and dynamics randomization \cite{tan2018sim} can be used.

\section{Conclusions}
In this paper, we present a hierarchical control framework for planning a sequence of learned latent actions from a high-level controller to a low-level policy. This framework allows us to accomplish several real-world locomotion tasks, such as goal-reaching, trajectory and velocity tracking with only 2000 samples on hardware. We present comparisons of our approach to other state-of-the-art hierarchical control approaches from literature -- including both learned and expert-designed approaches, on Daisy and A1 robot simulations. Our approach outperforms learning-based baselines on all tasks and performs similar to the expert-designed approach in normal settings while outperforming it when two hind legs are disabled in simulation. This work demonstrates the efficacy of a fully-learned hierarchical framework at achieving various locomotion tasks, and solving real-world problems like navigation on two very different robot platforms. It removes the need for user-designed action spaces, and opens up avenues for further research such as discovering new gaits, enabling sim-to-real transfer without performance loss, and navigating unknown terrains autonomously. 

\bibliographystyle{IEEEtran}
\bibliography{bibliography_bugfree}

\scriptsize{
\noindent \textbf{License for dataset used} Gibson Database of Spaces. License at \url{http://svl.stanford.edu/gibson2/assets/GDS_agreement.pdf}

}
\end{document}